\pdfoutput=1
\documentclass[11pt]{article}
\usepackage[]{acl}
\usepackage{times}
\usepackage{latexsym}
\usepackage[T1]{fontenc}
\usepackage[utf8]{inputenc}
\usepackage{latexsym}
\usepackage{array}
\usepackage{subfigure}
\usepackage{booktabs}
\usepackage{microtype}
\usepackage{url}
\usepackage{adjustbox}
\usepackage{adjustbox}
\usepackage{graphicx}
\usepackage{amssymb,amsmath,array}
\usepackage{multirow,makecell}
\usepackage{amsmath} 
\usepackage{amsfonts,amssymb}
\usepackage[normalem]{ulem}
\useunder{\uline}{\ul}{}

\title{Improving Agent Interactions in Virtual Environments with Language Models}

\author{Jack Zhang}

\begin{document}
\maketitle

\begin{abstract}
Enhancing AI systems with efficient communication skills for effective human assistance necessitates proactive initiatives from the system side to discern specific circumstances and interact aptly. This research focuses on a collective building assignment in the Minecraft dataset, employing language modeling to enhance task understanding through state-of-the-art methods. These models focus on grounding multi-modal understanding and task-oriented dialogue comprehension tasks, providing insights into their interpretative and responsive capabilities. Our experimental results showcase a substantial improvement over existing methods, indicating a promising direction for future research in this domain.

\end{abstract}

\section{Introduction}
The burgeoning field of artificial intelligence (AI) has introduced innovative approaches to collaborative tasks~\cite{jayannavar2020learning,nguyen2019help,roman2020rmm,madureira2023you,lachmy2021draw,narayan2019collaborative,shi-etal-2022-learning}, a notable example of which can be seen in the realm of construction. Within these tasks, the interplay between various roles, such as the builder and the architect, becomes crucial to the task's success.

In simpler terms, when people work together to build something, the person doing the building needs to really get what the plan maker wants. It's like playing a game of telephone where you have to be super clear and make sure you're on the same page. If something gets lost in translation, things can go wrong, showing just how important it is to talk things through and understand each other well.

This idea isn't just for building physical stuff; it also applies to how AI works. Here, the AI is like the builder, and it has to follow the instructions it gets (think of these as the blueprint) to make what's expected. The way the AI 'talks' and 'listens' to the instructions is key to making sure everything turns out right.

\begin{figure}[!t]
  \centering
  \includegraphics[width=0.5\textwidth]{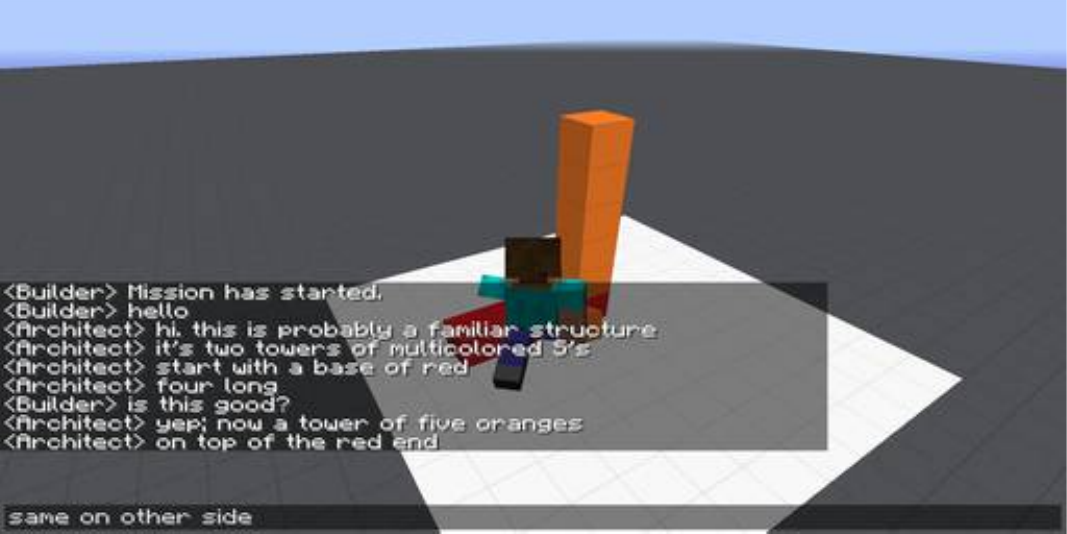}
  \caption{Within the ambit of a collaborative construction endeavor, it is incumbent upon the builder to adhere scrupulously to the directives issued by the architect. This endeavor mandates a thorough assimilation of the architect's specifications, as the culmination of the task hinges significantly on unambiguous communication and meticulous implementation. This framework accentuates the pivotal function of the builder in transmuting the architect's conceptualization into a concrete manifestation.
  }
  \label{fig:1}
\end{figure}

In previous studies, researchers tried to make an automatic building helper in the game Minecraft. This helper was supposed to figure out what to do and ask questions if something wasn't clear. But, these attempts didn't quite nail the part where the helper understands instructions really well and uses them correctly \cite{jayannavar2020learning,shi-etal-2022-learning}.

We're suggesting that using advanced language understanding techniques (like the technology behind smart assistants) could really help improve this.

In our study, we look at how well the AI, acting as the builder, follows instructions during a building task in Minecraft. We're checking if teaching the AI to understand language better helps it get the job done more effectively. Our findings are pretty exciting and suggest that AI could play a big role in working together with humans on projects.

\section{Related Works}
\paragraph{Language Modeling}
Research in language understanding, a core part of Natural Language Processing (NLP), has greatly changed and improved how computers understand human language. This progress moved from simple statistical approaches to today's more complex neural network models.

Initially, language models were based on statistics, like n-gram models mentioned by \cite{mikolov2013efficient}. Then came Word2Vec, also by \cite{mikolov2013efficient}, which was a game-changer by predicting words based on their context. Although these models were a step forward, they struggled with understanding longer text sequences and a wide range of vocabulary.

With the introduction of deep learning, models like RNNs, LSTMs, and GRUs \cite{pennington2014glove} began to improve how machines could grasp longer pieces of text. But the real leap came with transformer-based models, such as BERT \cite{devlin2018bert}. BERT, with its innovative approach to predicting missing parts of text and understanding the order of sentences, really changed the game. Research like RoBERTa by \cite{liu2019roberta} showed that some modifications could even boost performance in understanding sentences.

Further developments like ALBERT \cite{lan2019albert}, which aimed at better sentence-level predictions, and GPT by \cite{radford2019language}, focusing on predicting the next word in a sequence, showcased different strategies for enhancing language models. BART \cite{lewis2019bart} introduced another technique by mixing up sentences and then putting them back in order, helping models get better at seeing the big picture.

Recent research has continued to refine these models, focusing on training them for specific tasks to make them even more effective \cite{gururangan-etal-2020-dont,shi2023dont,alsentzer-etal-2019-publicly,shi2023rethinking}. This evolution underscores the ongoing efforts to make computers understand and use human language more like we do.

\begin{figure}[!t]
  \centering
  \includegraphics[width=0.5\textwidth]{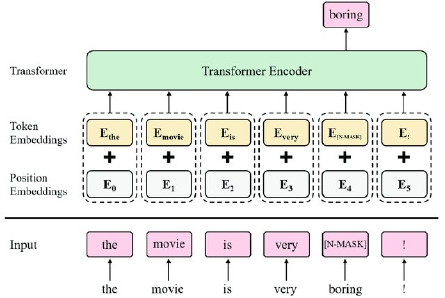}
  \caption{The example of masked language modeling \cite{devlin2018bert,liu2019roberta}.}
  \label{fig:mask}
\end{figure}

\paragraph{Language Grounding Tasks}
There's been a lot of interest from scientists in studying how machines can follow instructions through conversations. This has been explored in many research works, including tasks where machines navigate or find objects by asking questions and using visual hints, like in studies by \cite{suhr2018situated, suhr2019executing, chen2019touchdown, lachmy2021draw,de2017guesswhat,roman2020rmm,thomason2020vision}.

One interesting area is the Vision-and-Dialog Navigation (VDN) task, where the focus is on navigating with the help of visuals and dialogue, as discussed in \cite{chen2019touchdown, thomason2020vision,roman2020rmm, zhu2021self}. Tasks involving moving blocks \cite{misra2017mapping} or finding objects \cite{janner2018representation} are also part of this research trend.

A unique dataset called the Minecraft Corpus, introduced by \cite{narayan2019collaborative}, presents a task where an 'architect' works with a 'builder' in a game to create structures based on instructions. This was further developed by \cite{jayannavar2020learning} who created a model that follows these instructions step by step. Later, \cite{shi-etal-2022-learning} modified this into a task where the machine asks questions if it's confused about the instructions. This is similar to developments in VDN tasks, as noted in \cite{thomason2020vision,roman2020rmm,chi2020just}, where agents ask questions to clarify their next moves.

\cite{shi-etal-2022-learning} points out that understanding spatial relationships from text is a big challenge in these tasks. Many instruction-following tasks require grasping complex spatial and temporal ideas conveyed through language \cite{chen2019touchdown,yang2020robust,shi2022stepgame}. Achieving success in these tasks means having a deep understanding of natural language, which aligns well with our aim to improve language understanding through language modeling.

\paragraph{Pre-trained Language Models.} Pre-trained language models \cite{devlin2018bert,liu2019roberta,radford2019language,yamaguchi-etal-2021-frustratingly,alajrami-aletras-2022-pre} are at the center of natural language processing today. These Transformer based models are trained on large amounts of unlabeled data with language modeling objectives, and the resulting contextualized representations can be further fine-tuned on a wide range of downstream tasks.

Dynamic masking and static masking
where one masks $m$ (masking rate, typically 15\%) percentage of tokens from the original sentence $x$ and predicts on the masked token set $\mathcal{M}$. 

\paragraph{Self Training.} Self-training has historically been effective for NLP \cite{yarowsky-1995-unsupervised,mcclosky-etal-2006-effective}. Self-training learns about a task on a mix of labeled and unlabeled data. In self-training, the model learns as normal on labeled examples. On unlabeled examples, the model acts as both a teacher that makes predictions about the examples and a student that is trained on those predictions.

\section{Method}
In the next part, we'll talk about our new method for teaching an agent to build complicated designs from written instructions. This job is tough, especially when the agent needs to place items in certain spots on a grid.

\begin{figure}[!t]
  \centering
  \includegraphics[width=0.5\textwidth]{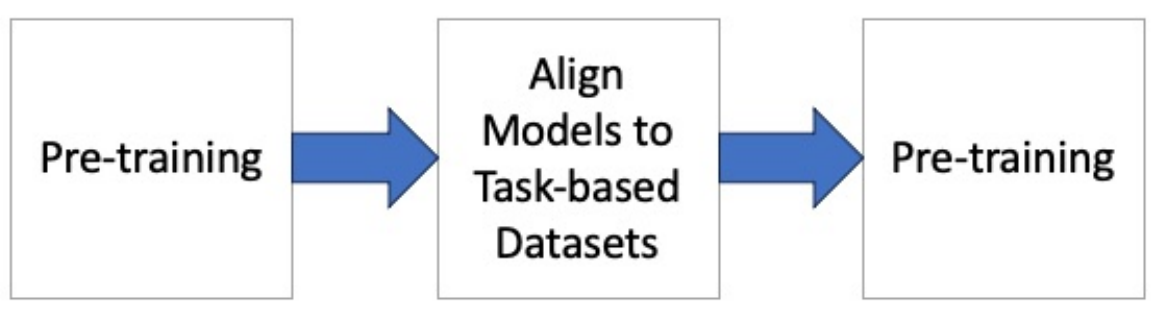}
  \caption{The flowchart of our method.}
  \label{fig:method}
\end{figure}

Our method is simple but really works well. We use a technique called masked language modeling on the text that describes what needs to be built. This way, we improve the language understanding of the agent, making it the main tool for the building tasks that follow, as shown in the figure referred to as \ref{fig:method}.

\begin{figure*}[!h]
    \centering
    \subfigure{\includegraphics[width=0.45\textwidth]{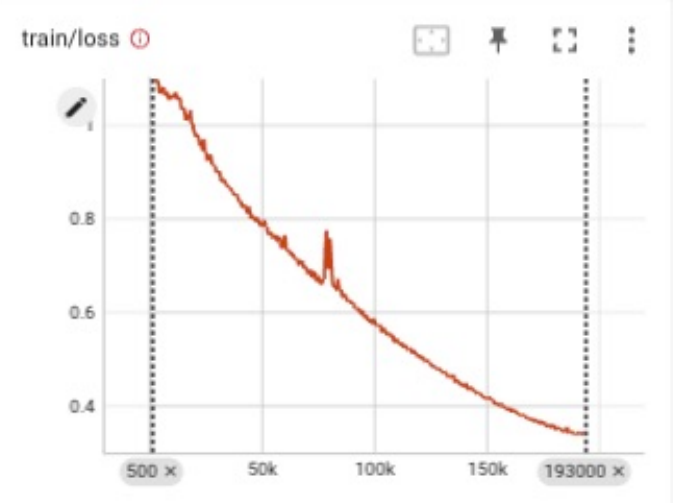}} 
    \subfigure{\includegraphics[width=0.45\textwidth]{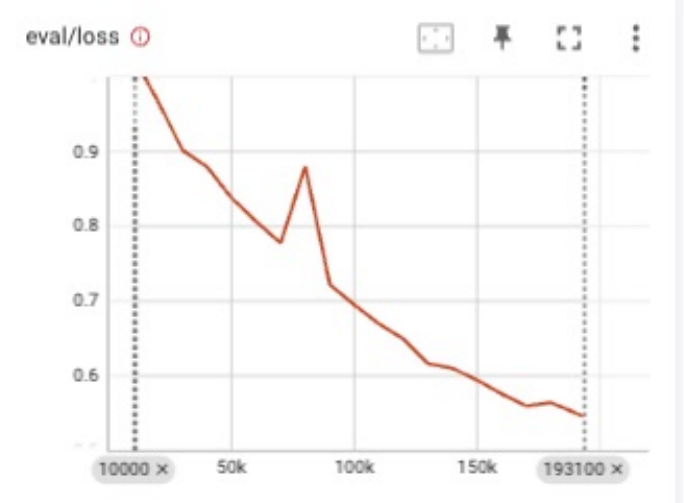}} 
    \caption{
    Experimental Result: Training and Validation Loss for masked language modeling.}
    \label{fig:loss}
\end{figure*}

\begin{figure}[h]
  \centering
  \includegraphics[width=0.5\textwidth]{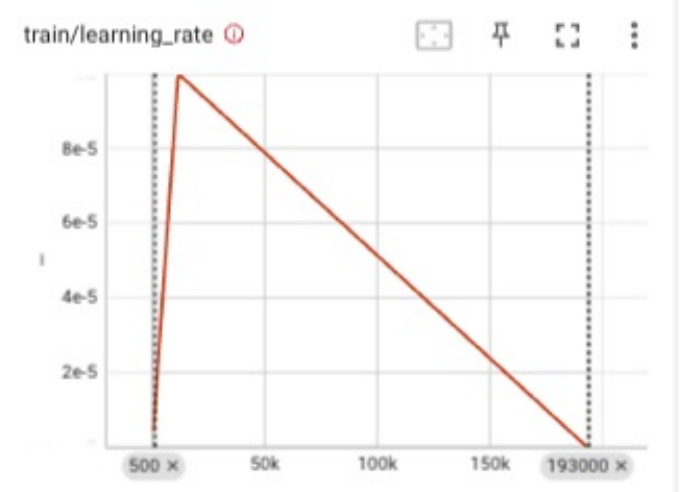}
  \caption{The change of learning rate during the training phrase.}
  \label{fig:lr}
\end{figure}

Specifically, it can be divided into two phrases:
\begin{itemize}
    \item 1. Choose a Pre-trained Language Model: Start by picking an existing pre-trained language model that fits your needs. This could be a popular choice like BERT \cite{devlin2018bert} or any other model that meets your criteria.;
    \item 2. Pick the Best Models for Your Task: Look for the top-performing models specifically designed for the task you're working on. These should be models that have set new standards for performance and are well-regarded in the research world. For instance, in our case, we chose the LearnToAsk model \cite{shi-etal-2022-learning}.;
    \item Adapt the Language Model to Your Task: Use the pre-trained language model you selected and apply it to your project's data. The aim here is to fine-tune the model so it gets better at understanding and producing language that's relevant to your specific task. After this, you'll end up with an updated version of the model.
    \item Use the Updated Model for Training: Finally, take this refined model and use it as the starting point to train the advanced models you identified earlier. This approach ensures these models benefit from the improved language capabilities developed in the previous step.
\end{itemize}

The main goal of the first step is to make our language models smarter \cite{liu2019roberta,devlin2018bert}, so they perform better when dealing with the specific job we want them to do. We do this by training these models on the details and challenges of the job at hand. This helps the models learn and get better at understanding and managing tasks like this in the future.

After training, these smarter models become the groundwork for the next set of tasks. With their upgraded ability to grasp and tackle the job's requirements, they offer a strong and effective foundation for these following tasks. The improved models are now better at processing, comprehending, and reacting to what's needed for these tasks, leading to more successful results.

\section{Experiment, Results and Discussion}

\begin{table*}[ht!]
\centering
\begin{tabular}{lccc}
\toprule
\bf Model & \bf Recall & \bf  Precision &  \bf F1  \\
\midrule
BAP model \cite{jayannavar2020learning}      & 12.6 & 22.4 & 16.1 \\
LearnToAsk \cite{shi-etal-2022-learning}    & 28.3 & 45.8 & 35.0 \\
Ours           & \bf28.5 & \bf46.3 & \bf35.3 \\
\bottomrule
\end{tabular}
\caption{Test Results for our proposed method and baseline models.}
\label{table:results}
\end{table*}

In this section, we detail the execution of our experiments and the analysis of the findings.

We established a clear and thorough framework for our experiments, aimed at rigorously evaluating the efficacy and efficiency of our proposed methodology across a variety of tasks. Each experiment was carried out under strictly controlled conditions to ensure the reliability of the results and to mitigate any possible biases or external interferences.

Following the completion of these experiments, we engaged in a meticulous examination and interpretation of the data collected. It became apparent that there was a marked enhancement in the language models' performance after undergoing our specialized training regimen. The evidence pointed to our method significantly boosting the models' proficiency in navigating the complexities associated with the downstream tasks.

Ultimately, the outcomes of our experimental endeavors lend strong support to the viability of our proposed strategy. These results substantiate the assertion that targeted, methodical training of language models on specific tasks can lead to substantial improvements in their performance on subsequent tasks, thereby affirming our original hypothesis and the effectiveness of our approach.

\paragraph{Dataset} For our experiments, we utilized the collaborative building datasets \cite{jayannavar2020learning,shi-etal-2022-learning}.

\paragraph{Baseline Models} Our method was benchmarked against two foundational models, namely the BAP model \cite{jayannavar2020learning} and the LearnToAsk Model.

\paragraph{Evaluation Metrics} The performance of the models was assessed using metrics such as the F1 score, Recall Rate, and Precision Rate. The F1 score, in particular, provided a comprehensive measure of accuracy by simultaneously taking into account both the precision and recall metrics.

\paragraph{Training Details} The training process was conducted using the cross-entropy loss function over a span of 100 epochs, specifically focusing on the masked language modeling task. This was complemented by a learning rate set at 1e-4. The baseline models underwent training as per the specifications laid out in their original publications.

\paragraph{Results} In Table \ref{table:results}, we present a comparative analysis of the performance of our proposed model against the baseline models on the Minecraft Corpus Dataset, particularly in relation to the collaborative building task. The data clearly shows our model outperforming the baselines, underscoring the success and impact of our proposed method.

\paragraph{Analysis of Loss Metrics} Figure \ref{fig:loss} provides a visual representation of the training and validation loss metrics over the course of the training period. A consistent downward trend in these metrics is observed, indicative of the model's ongoing learning and improvement in its predictive capabilities. This decline in loss values signifies the model's enhanced generalization ability, optimizing its performance through successive iterations over the dataset.

\paragraph{Learning Rate Evolution} Figure \ref{fig:lr} depicts the progression of the learning rate during the masked language modeling task. Monitoring the adjustments in the learning rate is essential as it plays a pivotal role in determining the training pace and the overall effectiveness of the model's learning process. Through this analysis, we are afforded a deeper understanding of the model's adaptive learning strategies and the pace at which it optimizes its performance throughout the training phase.

\section{Conclusion}
In summary, our research highlights the importance of advanced AI communication abilities and the initiative of AI systems to match human comprehension and offer effective support in diverse scenarios.

By investigating a cooperative construction task using the Minecraft dataset, we have introduced an innovative approach that utilizes language modeling to deepen task comprehension. The use of cutting-edge models has markedly boosted understanding across different modes of communication and dialogue that's focused on tasks. The findings from our experiments support the effectiveness of our approach, indicating a significant improvement in how well these state-of-the-art models perform in complicated tasks.

\bibliography{thesis}
\bibliographystyle{acl_natbib}

\end{document}